%% file: main.tex
\documentclass[sigconf]{acmart}
\setkeys{acmart.cls}{balance=false}

\AtBeginDocument{%
  \providecommand\BibTeX{{%
    \normalfont B\kern-0.5em{\scshape i\kern-0.25em b}\kern-0.8em\TeX}}}

\setcopyright{none}

\copyrightyear{2027}
\acmYear{2027}
\setcopyright{cc}
\setcctype{by}
\acmConference[ASPDAC '27]{32nd Asia and South Pacific Design Automation Conference}{January 25--28, 2027}{Tokyo, Japan}
\acmBooktitle{32nd Asia and South Pacific Design Automation Conference (ASPDAC '27), January 25--28, 2027, Tokyo, Japan}
\acmDOI{10.1145/3840404.3860924}
\acmISBN{979-8-4007-2923-2/2027/01}

\input{package.tex}

\begin{document}




\title{Self-Improving Agents for RTL PPA Optimization: \\ Benchmarks and Tool-Grounded Optimization}
\title{Toward Autonomous RTL and Synthesis Optimization: Benchmarks, Self-Improving Agents, and Algorithm Discovery}
\title{Toward Autonomous RTL and Synthesis Optimization: Benchmarking and Self-Improving Agents}
\subtitle{Invited Paper}


\author{Wenji Fang$^\dagger$, Yao Lu$^\dagger$, Jinghua Wang, Zhiyao Xie$^\star$}
\affiliation{%
  \institution{Hong Kong University of Science and Technology}
  \country{}
}

\input{section/abstract.tex}

\maketitle

\begingroup\renewcommand\thefootnote{$\dagger$}
\footnotetext{Equal Contribution}
\endgroup

\begingroup\renewcommand\thefootnote{$*$}
\footnotetext{Corresponding Author (eezhiyao@ust.hk)}
\endgroup
\thispagestyle{empty}

\pagestyle{empty}

\input{section/1-intro.tex}

\input{section/2-benchmark.tex}

\input{section/2-preliminaries.tex}
\input{section/3-method.tex}

\input{section/3-synthesis.tex}

\input{section/4-experiments.tex}
\input{section/5-discussion.tex}

\input{section/6-conclusion}

\clearpage
\bibliographystyle{ACM-Reference-Format}
\bibliography{ref}


\end{document}

%% file: package.tex
\usepackage{hhline}
\usepackage{colortbl}
\usepackage{centernot}
\usepackage{pbox}
\usepackage{amsmath}
\usepackage{amsfonts}
\usepackage{url}
\usepackage{bm}
\usepackage{comment}
\usepackage{graphicx}
\usepackage{subfigure}
\usepackage{epstopdf}
\usepackage{multirow}
\usepackage{color}
\usepackage{tabu}
\usepackage{mdframed}
\usepackage{syntax}
\usepackage{fancyvrb}
\usepackage{cleveref}

\usepackage[ruled,linesnumbered,algo2e]{algorithm2e}
\usepackage[noend]{algpseudocode}
\usepackage{algorithmicx}
\usepackage{algpseudocode}
\usepackage{paralist}
\usepackage{stmaryrd}
\usepackage{tikz}
\usepackage[referable]{threeparttablex}
\usepackage{tabularx}
\usepackage{booktabs}
\usepackage{array}
\usepackage{fancyhdr}
\usepackage{float}
\usepackage{xspace}
\usepackage{pifont}
\usepackage{balance}
\usepackage{xcolor}

\usepackage{tablefootnote}
\usepackage[normalem]{ulem}
\usepackage{bbding}

\usepackage{tikz}
\usepackage{xcolor}

\usepackage{graphicx}
\usepackage{adjustbox}

\usepackage{float}

\newfloat{figtab}{htb}{fgtb}
\makeatletter
  \newcommand\figcaption{\def\@captype{figure}\caption}
  \newcommand\tabcaption{\def\@captype{table}\caption}
\makeatother

\definecolor{princetonorange}{RGB}{255,143,0}

\definecolor{lightgreen}{RGB}{198, 224, 183}

\definecolor{lightred}{RGB}{240, 205, 176}

\definecolor{agentrtl}{HTML}{CDE09A}

\usepackage{geometry}
\definecolor{USTgold}{RGB}{153,102,0}
\definecolor{USTyellow}{RGB}{204,153,0}
\definecolor{USTyellowlight}{RGB}{255,212,0}
\definecolor{USTorange}{RGB}{255,166,26}
\definecolor{USTpink}{RGB}{255,157,157}
\definecolor{USTblue}{RGB}{0,51,102}
\definecolor{USTmiddle}{RGB}{0,116,188}
\definecolor{USTlight}{RGB}{99,202,225}
\definecolor{USTgray}{RGB}{204,204,204}
\definecolor{USTred}{RGB}{237,27,47}
\definecolor{USTdarkred}{RGB}{124,35,72}

\usepackage{tcolorbox}
\tcbuselibrary{skins,breakable}
    {\endtcolorbox}

%% file: section/abstract.tex
\begin{abstract}
Recent advances in large language models (LLMs) and autonomous agents
are creating new opportunities for hardware design optimization.
This paper examines three connected directions toward autonomous RTL
and synthesis optimization:
(1) \emph{Realistic benchmarking.}
We discuss reference-based and open-ended RTL optimization benchmarks
that evaluate optimization opportunities under realistic EDA workflows.
(2) \emph{Tool-grounded RTL optimization.}
We present Dr.~RTL, a multi-agent framework that automates
critical-path analysis, RTL rewriting, and implementation-level
evaluation, while continuously improving through reusable optimization
skills. Evaluated on 20 real-world RTL designs, Dr.~RTL achieves
average WNS/TNS improvements of 21\%/17\% with a 6\% area reduction
over commercial synthesis.
(3) \emph{Design-specific synthesis optimization.}
We explore agent-driven synthesis script tuning and the discovery of
executable synthesis algorithms across intermediate representations,
technology-independent optimization, and technology mapping.
Finally, we envision AI-native design automation that jointly explores
architectural, RTL, and EDA algorithm design spaces, guided by
verification constraints and implementation feedback.
\end{abstract}

%% file: section/1-intro.tex
\section{Introduction}
\label{sec:intro}

\textbf{RTL optimization and logic synthesis are two
synergistic processes that jointly determine the power,
performance, and area (PPA) of digital circuits.}
Given a hardware architecture, RTL optimization explores
alternative implementations through logic restructuring,
resource sharing, and sequential transformations, while
synthesis and place-and-route (P\&R) tools further optimize
the circuit through logic synthesis and physical
implementation~\cite{vahid2010digital,hassoun2012logic}.
The two optimization processes are highly synergistic:
RTL structure determines the optimization opportunities
available to downstream tools, while synthesis and
physical design feedback reveals circuit bottlenecks
that can guide further RTL transformations.
Achieving better PPA therefore requires considering
RTL optimization and downstream EDA optimization
together through an iterative, tool-grounded process.

\textbf{This synergy also raises a fundamental question:
\emph{Which RTL optimization opportunities remain
after synthesis optimizations?}}
Modern synthesis tools can automatically simplify,
restructure, and map complex logic, potentially
absorbing many source-level RTL transformations.
Consequently, an RTL rewrite that improves PPA under
a weak synthesis flow may offer little additional
benefit under a stronger commercial tool.
However, existing RTL optimization studies often
evaluate manually degraded designs using open-source
synthesis flows, as in RTLRewriter~\cite{yao2024rtlrewriter}
and SymRTLo~\cite{wang2025symrtlo}.
Such evaluations may emphasize recovering injected
inefficiencies or transformations already handled
by advanced synthesis.
More meaningful RTL optimization should target
opportunities that remain beyond realistic synthesis,
while improvements to synthesis algorithms themselves
can further extend the achievable PPA.

\begin{figure}[!t]
    \centering
    \includegraphics[width=\linewidth]{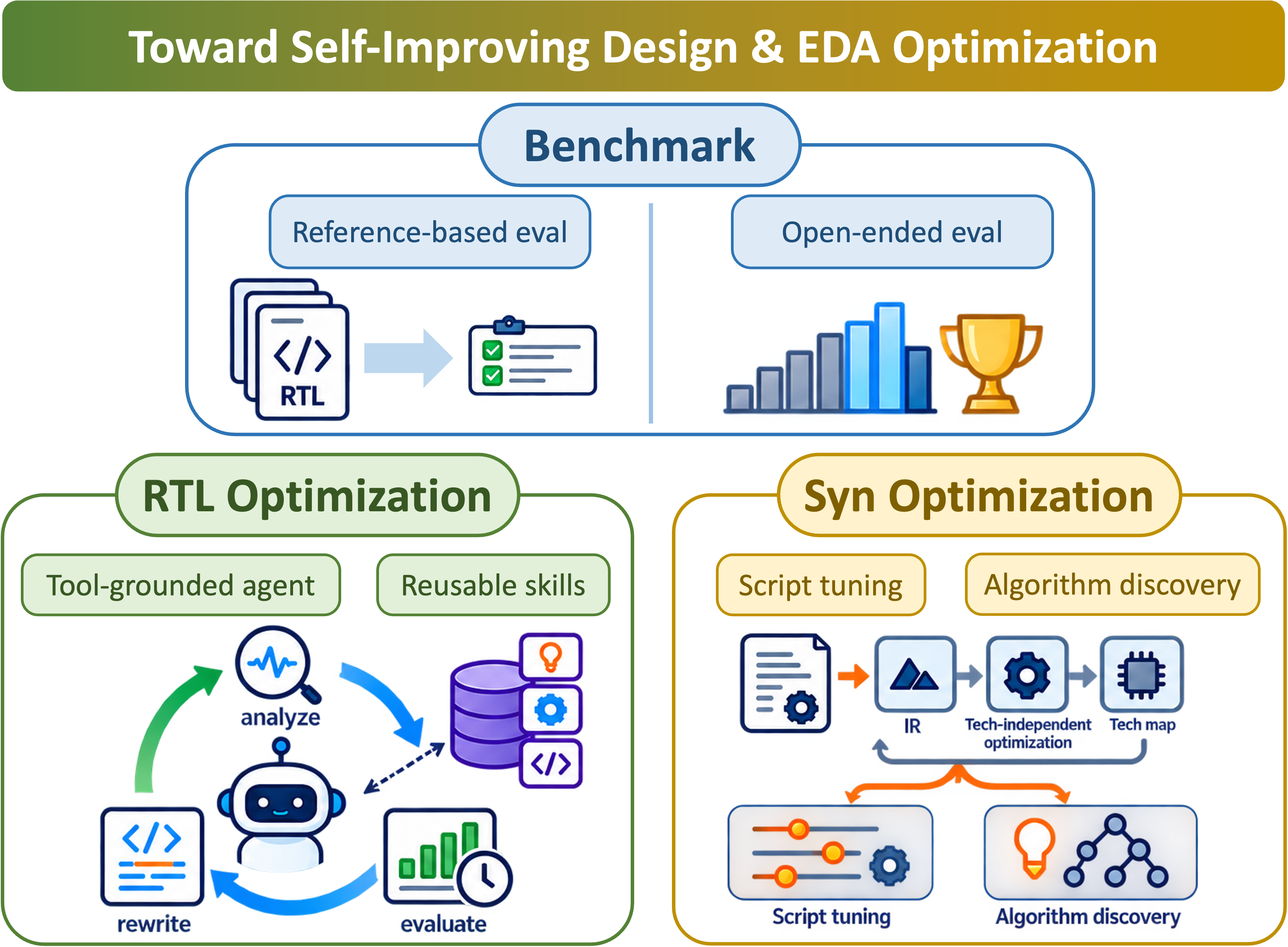}
    \vspace{-.25in}
    \caption{Three connected directions toward autonomous
    RTL and synthesis optimization. Realistic benchmarking
    establishes evaluation settings, tool-grounded agents
    optimize RTL beyond synthesis, and synthesis optimization
    extends agentic search from flow tuning to
    algorithm discovery.}
    \label{fig:overview}
    \vspace{-.1in}
\end{figure}

This perspective motivates two fundamental research
challenges toward autonomous RTL and synthesis optimization:

\begin{itemize}
    \item \textbf{How to benchmark RTL optimization?}
    Establish realistic RTL designs and evaluation
    settings that distinguish genuine optimization
    opportunities from improvements already achievable
    through synthesis.

    \item \textbf{How to automate and improve optimization?}
    Develop tool-grounded agents that explore RTL
    implementations and synthesis strategies,
    while accumulating reusable optimization knowledge.
\end{itemize}

These challenges connect three research directions:
realistic RTL optimization benchmarking, agentic RTL
optimization, and synthesis tool optimization.
We briefly review their recent progress and
remaining challenges below.

\textbf{Benchmarking realistic RTL optimization.}
Recent advances in LLM-based RTL generation~\cite{fang2025survey} include
domain-specific models~\cite{liu2024rtlcoder},
verification-guided generation~\cite{sami2024aivril},
and agentic frameworks~\cite{zhao2025mage,deng2025scalertl}.
Corresponding benchmarks, such as
VerilogEval~\cite{liu2023verilogeval},
RTLLM~\cite{lu2024rtllm},
OpenLLM-RTL~\cite{liu2024openllmrtl},
and CVDP~\cite{pinckney2025comprehensive},
primarily evaluate functional correctness.
RTL optimization additionally requires measuring
implementation quality after synthesis while
preserving functional equivalence.
Existing evaluations often rely on manually degraded
RTL and relatively weak synthesis flows, leaving
a gap between benchmark results and optimization
under industrial conditions.
This motivates realistic benchmarks that evaluate
both known optimization targets and open-ended
improvements on original human-written RTL,
with synthesis quality and verification incorporated
into the evaluation.

\textbf{Tool-grounded RTL optimization.}
With realistic synthesis evaluation, the next challenge
is to discover RTL transformations that provide
additional PPA improvements.
Their effectiveness depends on how downstream tools
restructure and map the resulting logic, making
implementation feedback essential.
Existing approaches explore verified rewriting
~\cite{coward2024rover},
LLM-based RTL optimization~\cite{yao2024rtlrewriter},
symbolic reasoning~\cite{wang2025symrtlo},
timing-oriented transformations~\cite{xu2025rethinking},
evolutionary optimization~\cite{ping2026poet,hsin2026evolve},
and multi-agent collaboration~\cite{chang2026codmas}.
Related studies investigate retrieval-guided circuit
optimization~\cite{chowdhury2024retrieval}
and PPA-aware RTL generation
~\cite{thorat2025llm,tasnia2025veriopt}.
Fine-grained timing analysis~\cite{fang2024annotating}
can reveal bottlenecks that remain after synthesis,
while equivalence checking ensures correctness.
These capabilities motivate self-improving agents
that use detailed EDA feedback to explore RTL
transformations and learn reusable optimization strategies.

\textbf{Synthesis tool optimization.}
The synergy between RTL and synthesis also suggests
another optimization direction: improving the
synthesis procedures themselves.
Existing synthesis techniques explore Boolean
rewriting~\cite{brayton2006scalable},
technology-aware optimization~\cite{chen2024syn},
and symbolic equality saturation~\cite{yin2025boole}.
Automated flow tuning further explores synthesis
parameters~\cite{ziegler2016synthesis},
configuration selection~\cite{xie2020fist},
and optimization sequences~\cite{neto2022flowtune}.
More recently, AlphaEvolve~\cite{novikov2025alphaevolve}
has demonstrated the feasibility of AI-driven
algorithm discovery, while
Self-Evolved ABC~\cite{yu2026selfevolvedabc}
applies code evolution to improve general-purpose
logic synthesis.
These advances motivate design-specific synthesis
optimization, where agents can tune commercial
synthesis flows and discover customized algorithms
that better exploit individual circuit structures.
Together with RTL optimization, this opens
opportunities for more comprehensive RTL--EDA
co-optimization.

In this paper, we review three connected research
directions toward autonomous RTL and synthesis
optimization and present our recent progress
in each direction, as illustrated in~\Cref{fig:overview}.
Our contributions are summarized as follows:

\begin{itemize}
    \item \textbf{Realistic benchmarking of RTL optimization.}
    We introduce two complementary benchmarking settings:
    reference-based optimization through RTL-OPT and
    open-ended optimization on original human-written
    RTL through Dr.~RTL.
    Together, they evaluate both known optimization
    opportunities and further PPA improvements
    under realistic synthesis and formal
    verification workflows.

    \item \textbf{Tool-grounded, self-improving agentic
    RTL optimization.}
    We present Dr.~RTL, a multi-agent framework
    that combines fine-grained timing analysis,
    parallel RTL exploration, sequential equivalence
    checking, and group-relative skill learning.
    Across 20 human-written RTL designs, Dr.~RTL
    achieves average WNS and TNS improvements of
    21.3\% and 16.9\%, respectively, with a 5.8\%
    reduction in area over commercial synthesis.

    \item \textbf{From synthesis flow tuning to
    design-specific algorithm discovery.}
    We explore two levels of synthesis optimization:
    agent-driven script tuning for commercial synthesis
    tools and design-specific algorithm discovery
    for open-source synthesis tools.
    Across eight designs, commercial synthesis tuning
    improves average WNS/TNS by 7.4\%/11.1\%,
    while open-source algorithm discovery reduces
    area-delay product (ADP) by 18.7\%.
\end{itemize}

\textbf{Looking ahead, we envision a self-improving
AI-driven design automation paradigm built on
verification-constrained optimization spaces.}
These spaces span three levels: architecture,
RTL implementation, and EDA algorithms.
At the architectural level, agents explore
microarchitectures, dataflows, and resource
organizations.
At the RTL level, they optimize circuit
implementations, while at the EDA tool level,
they improve synthesis and physical design
procedures.
Verification defines what must be preserved
at each level, establishing the boundaries
for optimization.
With these spaces and constraints clearly
defined, a key research direction is to develop
more effective agentic design-space exploration
across abstraction levels.
Accumulated optimization strategies and
verification knowledge can support continual
self-improvement and the joint optimization
of hardware implementations and EDA tools.

%% file: section/2-benchmark.tex
\section{Benchmarking RTL Optimization}
\label{sec:benchmark}

\subsection{Overview}
\label{sec:benchmark-overview}

RTL optimization benchmarks must evaluate both functional
correctness and implementation quality after synthesis.
Since the PPA benefit of RTL transformations depends on
downstream synthesis, benchmark quality is determined by
the starting RTL designs, synthesis toolchain, verification
flow, and evaluation metrics.
In particular, a realistic benchmark should capture
optimization opportunities that remain meaningful under
strong synthesis tools.

\begin{figure}[!t]
    \centering
    \includegraphics[width=\linewidth]{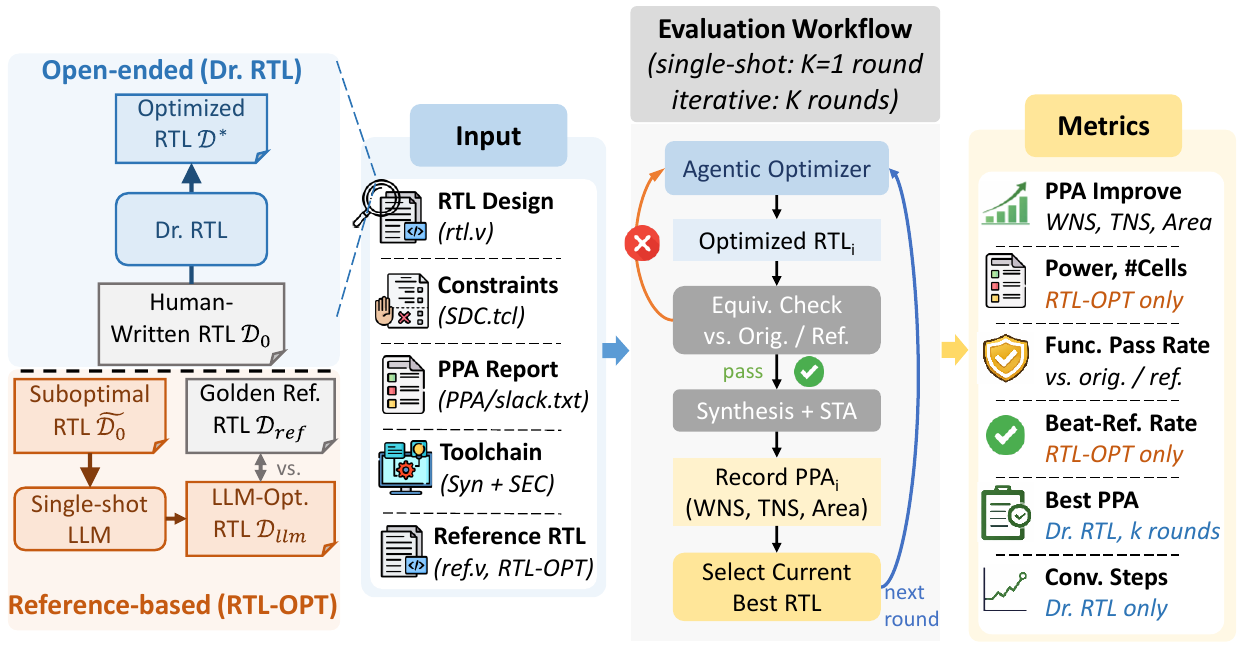}
    \vspace{-.25in}
    \caption{Two complementary RTL optimization evaluation
    settings. Reference-based evaluation measures improvements
    against human-optimized RTL references, while open-ended
    evaluation measures further PPA improvements on original
    human-written designs. Both settings assess implementation
    quality after synthesis and functional verification.}
    \label{fig:benchmark}
    \vspace{-.1in}
\end{figure}

We introduce two complementary evaluation settings,
as illustrated in~\Cref{fig:benchmark}.
\emph{Reference-based evaluation} starts from suboptimal RTL
and measures optimization against human-optimized references.
\emph{Open-ended evaluation} starts from original human-written
RTL and measures further improvements beyond commercial
synthesis.
The former provides known optimization targets, while
the latter evaluates the ability to discover additional
optimization opportunities without predefined references.

\subsection{Background and Related Work}
\label{sec:benchmark-related}

\textbf{RTL generation benchmarks.}
LLM-based hardware design has motivated several benchmarks
for evaluating RTL generation from natural-language
specifications, such as
VerilogEval~\cite{liu2023verilogeval},
RTLLM~\cite{lu2024rtllm}, and
OpenLLM-RTL~\cite{liu2024openllmrtl}.
These benchmarks primarily evaluate syntactic and
functional correctness.
RTL optimization additionally requires measuring
PPA improvements while preserving the original
design functionality.

\textbf{RTL optimization evaluations.}
Existing RTL optimization methods are evaluated under
different design and toolchain settings.
RTLRewriter~\cite{yao2024rtlrewriter}
and SymRTLo~\cite{wang2025symrtlo}
explore RTL rewriting using their respective synthesis
and verification workflows.
POET~\cite{ping2026poet} investigates power-oriented
optimization, while CODMAS~\cite{chang2026codmas}
uses multi-agent collaboration for RTL optimization.
Differences in starting designs, optimization objectives,
and synthesis tools make their results difficult
to compare directly.

Many existing evaluations use manually degraded RTL,
where optimization aims to recover injected inefficiencies.
Such settings provide controlled tasks, but the measured
gains depend on how degradation is introduced and how
effectively synthesis handles the resulting structures.
This motivates complementary benchmarks that evaluate
both known optimization gaps and open-ended improvement
opportunities under realistic synthesis conditions.

\subsection{Reference-Based RTL Optimization: RTL-OPT}
\label{sec:rtlopt}

\subsubsection{Benchmark Description}
\label{sec:rtlopt-desc}

RTL-OPT~\cite{lu2026new} introduces a reference-based
benchmark using paired suboptimal and designer-optimized
RTL implementations.
Each task provides a functionally correct but suboptimal
RTL design and a corresponding human-optimized reference.
The benchmark contains 36 handcrafted optimization tasks
covering arithmetic logic, control logic, finite-state
machines, and pipelined datapaths.
These tasks are designed to expose RTL-level optimization
opportunities that persist after synthesis.

\begin{table}[!t]
\caption{Comparison of LLM-based RTL optimization under
(a) reference-based and (b) open-ended evaluation settings.}
\label{tbl:baseline2}
\vspace{-.15in}
\centering
\resizebox{0.45\textwidth}{!}{
\begin{tabular}{c|c||ccc}
\toprule
\multicolumn{5}{c}{
    \textit{(a) Reference-based: RTL-OPT}
} \\
\midrule
\multicolumn{2}{c||}{\textbf{Method}}
    & Func. Pass & $>$Subopt. & $>$Ref. \\
\midrule
\multirow{4}{*}{
    \begin{tabular}[c]{@{}c@{}}
        Single-Shot\\LLM
    \end{tabular}
}
    & GPT-4o-mini & 75.0\% & 19.4\% & 5.6\% \\
    & Gemini-2.5  & 69.4\% & 30.6\% & 5.6\% \\
    & DeepSeek V3 & 69.4\% & 27.8\% & 13.9\% \\
    & DeepSeek R1 & 61.1\% & 41.7\% & 13.9\% \\
\toprule
\multicolumn{5}{c}{
    \textit{(b) Open-ended: Original RTL}
} \\
\midrule
\multicolumn{2}{c||}{\textbf{Method}}
    & WNS & TNS & Area \\
\midrule
\multirow{2}{*}{
    \begin{tabular}[c]{@{}c@{}}
        Single-Shot\\LLM
    \end{tabular}
}
    & Claude Opus & -2.4\% & -2.8\% & -0.7\% \\
    & GPT 5.3     & -1.2\% & 0.3\%  & 0.6\% \\
\midrule
\multirow{2}{*}{
    \begin{tabular}[c]{@{}c@{}}
        SOTA\\Baselines$^\star$
    \end{tabular}
}
    & RTLRewriter~\cite{yao2024rtlrewriter}
        & -4.9\% & -6.3\% & -3.1\% \\
    & SymRTLo~\cite{wang2025symrtlo}
        & -7.1\% & -5.7\% & -1.4\% \\
\midrule
\rowcolor{agentrtl}
\textbf{Ours}
    & \textbf{Dr.~RTL}
    & \textbf{-21.3\%}
    & \textbf{-16.9\%}
    & \textbf{-5.8\%} \\
\bottomrule
\end{tabular}
}
\begin{tablenotes}\footnotesize
\item $^\star$ The iterative RTL optimization baselines
use the same backbone LLM, Claude Opus, as Dr.~RTL.
$>$Subopt. and $>$Ref. indicate the fractions of tasks
whose PPA exceeds the suboptimal input and the
human-optimized reference, respectively.
In the open-ended setting, negative percentage changes
indicate reductions in timing violations or area.
\end{tablenotes}
\vspace{-.2in}
\end{table}

\subsubsection{Evaluation Metrics}
\label{sec:rtlopt-metrics}
 
RTL-OPT evaluates functional correctness and post-synthesis
PPA quality.
Candidate implementations are checked for functional
equivalence and synthesized to measure power, timing,
and area, including total power, worst negative slack (WNS),
total negative slack (TNS), cell area, and cell count.
 
Optimization capability is summarized by two fractions over all tasks. The first is the fraction of candidates whose PPA outperforms the suboptimal input, and the second is the fraction whose PPA outperforms the human-optimized reference. The individual PPA metrics are still reported alongside these fractions, so that the trade-offs among different optimization objectives are preserved.
\subsubsection{Evaluation Results}
\label{sec:rtlopt-results}

Under commercial \texttt{compile\_ultra} synthesis,
35 of the 36 human-optimized references outperform
their corresponding suboptimal implementations,
confirming that the benchmark retains meaningful
optimization opportunities after synthesis.
As shown in~\Cref{tbl:baseline2}, existing LLMs
show limited ability to recover these improvements.
The highest fraction of tasks outperforming the
suboptimal input is 41.7\%, while only 13.9\%
exceed the human-optimized reference.
These results highlight the difficulty of finding
RTL transformations that translate into better
implementations under commercial synthesis.

\subsection{Open-Ended RTL Optimization Evaluation}
\label{sec:drrtl}

\subsubsection{Benchmark Description}
\label{sec:drrtl-desc}

Open-ended evaluation measures optimization opportunities
remaining in original human-written RTL implementations,
without assuming a known optimized reference.
This setting is introduced and adopted in
Dr.~RTL~\cite{xxx}, using 20 designs collected from
diverse open-source projects, including processors,
communication interfaces, buffers, and signal-processing
modules.
The designs average 812 lines of code and three modules,
with more complex structures and hierarchies than
typical small-scale RTL optimization tasks.

The evaluation uses commercial logic synthesis for implementation quality and sequential equivalence checking (SEC) for functional correctness. 
Starting from human-written RTL provides a more challenging
optimization setting, where improvements must go beyond
those already achieved by the synthesis tool.

\begin{figure*}[!t]
    \centering
    \vspace{-.3in}
    \includegraphics[width=\linewidth]{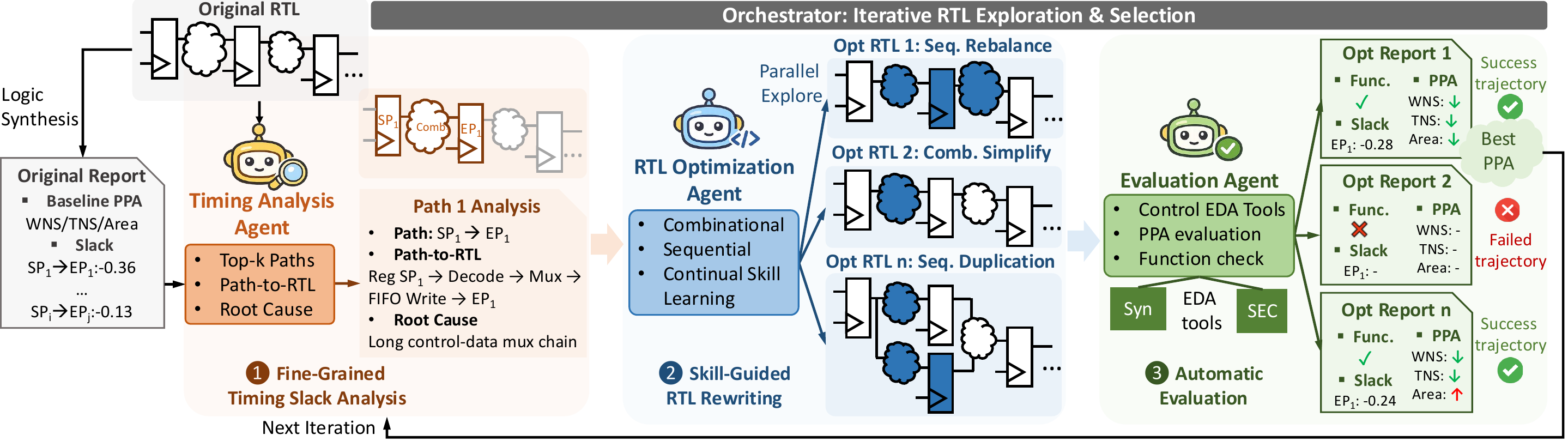}
    \vspace{-.25in}
    \caption{Overview of Dr.~RTL. The framework coordinates
    critical-path analysis, parallel RTL rewriting, and
    tool-based evaluation. Commercial synthesis provides
    implementation feedback, while sequential equivalence
    checking verifies functional correctness.}
    \vspace{-.2in}
    \label{fig:agent}
\end{figure*}

\begin{figure}[!t]
    \centering
    \includegraphics[width=\linewidth]{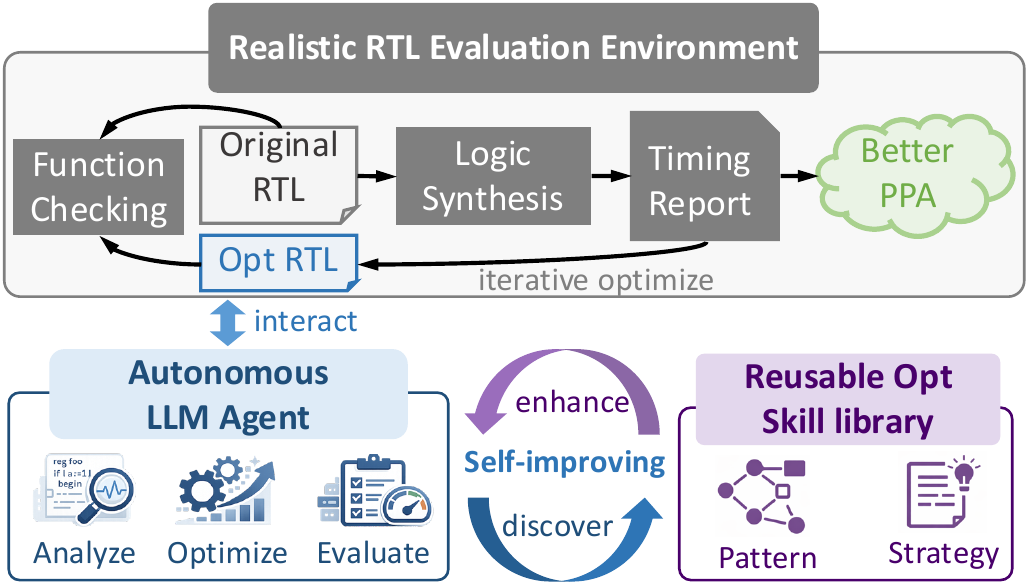}
    \vspace{-0.25in}
    \caption{Dr.~RTL automates RTL PPA optimization through
    closed-loop interaction with EDA tools and distills
    optimization trajectories into reusable skills
    for continual self-improvement.}
    \label{fig:motivation}
    \vspace{-.25in}
\end{figure}

\subsubsection{Evaluation Metrics}
\label{sec:drrtl-metrics}

The benchmark measures PPA improvements relative to
the original RTL after commercial synthesis.
Timing is evaluated using WNS and TNS, while
implementation cost is measured using cell area.
SEC verifies that candidate implementations preserve
the original sequential behavior.
Power is excluded from the current optimization objective
because reliable power estimation requires
workload-dependent switching activity.

Beyond final PPA, the open-ended setting also supports evaluating iterative optimization through functional pass rates and optimization trajectories. These metrics characterize both the quality of the optimized implementations and the reliability of the optimization methods that produce them.

\subsubsection{Evaluation Results}
\label{sec:drrtl-results}

As shown in~\Cref{tbl:baseline2}, open-ended evaluation
reveals substantial differences in optimization capability
under commercial synthesis.
Single-shot LLMs provide limited improvements on
original human-written RTL, while representative
iterative baselines achieve average WNS improvements
of 4.9\%--7.1\%.
In comparison, the Dr.~RTL optimizer achieves average
WNS and TNS improvements of 21.3\% and 16.9\%,
respectively, with a 5.8\% reduction in area.

These results demonstrate that meaningful RTL
optimization opportunities remain beyond commercial
synthesis, while their discovery requires effective
optimization methods.
The Dr.~RTL framework is detailed in~\Cref{sec:method}.

%% file: section/3-method.tex
\section{Agentic RTL Optimization with Dr.~RTL}
\label{sec:method}

\subsection{Overview}

Dr.~RTL~\cite{xxx} is a tool-grounded, self-improving agentic
framework for optimizing RTL implementations beyond commercial
synthesis.
As illustrated in~\Cref{fig:motivation}, it automates the
iterative optimization process through interaction with EDA tools
and continuously accumulates reusable optimization knowledge.

\begin{figure*}[!t]
    \centering
    \vspace{-.3in}
    \includegraphics[width=\linewidth]{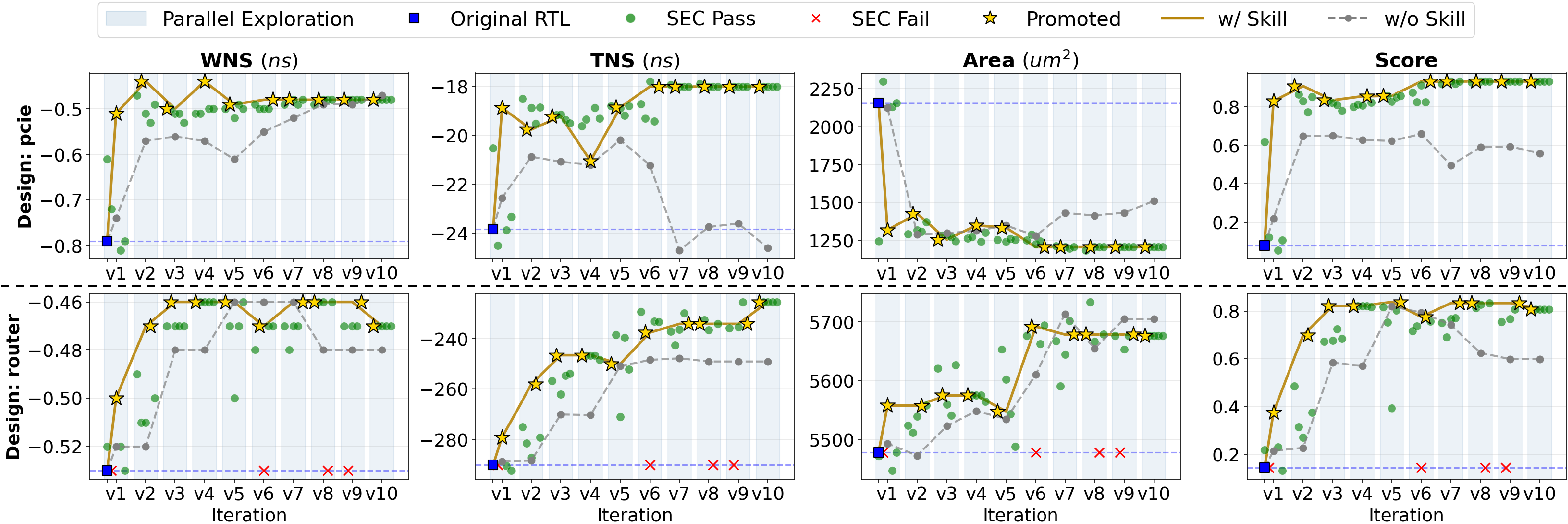}
    \vspace{-.25in}
    \caption{Optimization trajectories of Dr.~RTL across
    iterations for two designs (router and pcie).
    WNS, TNS, area, and optimization score improve
    over successive iterations while preserving
    functional equivalence. The skill-enhanced variant
    converges faster and achieves better PPA than
    the variant without learned skills.}
    \label{fig:trajectory}
    \vspace{-0.1in}
\end{figure*}

The framework combines two complementary components.
First, a multi-agent optimization workflow performs fine-grained
critical-path analysis, parallel RTL rewriting, and tool-based
evaluation under sequential equivalence checking (SEC).
Second, group-relative skill learning distills successful and
unsuccessful optimization experiences into reusable
pattern--strategy skills.
Together, these components enable agents to explore
functionally equivalent RTL implementations, learn from
implementation feedback, and improve subsequent searches
without updating the LLM parameters.

\subsection{Background and Related Work}

RTL optimization explores functionally equivalent
implementations for better post-synthesis PPA.
Its effectiveness depends on downstream synthesis,
requiring tool-grounded evaluation and equivalence checking.

\textbf{RTL rewriting and optimization.}
Existing approaches include verified e-graph rewriting
in ROVER~\cite{coward2024rover},
LLM-guided rewriting in RTLRewriter~\cite{yao2024rtlrewriter},
symbolic reasoning in SymRTLo~\cite{wang2025symrtlo},
and timing-oriented transformations~\cite{xu2025rethinking}.
Related studies explore retrieval-guided circuit
optimization~\cite{chowdhury2024retrieval}
and PPA-aware RTL generation
~\cite{thorat2023advanced,thorat2025llm,tasnia2025veriopt}.

\textbf{Agentic RTL optimization.}
Recent methods employ multi-agent collaboration
in CODMAS~\cite{chang2026codmas},
power-oriented evolutionary search in POET~\cite{ping2026poet},
and LLM-driven evolution in EvolVE~\cite{hsin2026evolve}.
Agentic RTL generation further explores multi-agent
design~\cite{zhao2025mage},
context evolution~\cite{deng2026ace},
and reinforcement learning~\cite{zhang2026rtlseek}.

\textbf{Tool feedback and reusable knowledge.}
Fine-grained RTL timing analysis~\cite{fang2024annotating}
localizes implementation bottlenecks, while sequential
equivalence checking~\cite{SEC} ensures correctness.
Agentic memory~\cite{xu2025mem} and skill learning
~\cite{xia2026skillrl} enable experience reuse.
Dr.~RTL combines critical-path feedback, commercial
synthesis, and group-relative skill learning for
self-improving RTL optimization.

\subsection{Method: Tool-Grounded Self-Improving RTL Optimization}
\label{sec:method1}


\textbf{Automating RTL optimization.}
As shown in~\Cref{fig:agent}, Dr.~RTL formulates
RTL optimization as an iterative, tool-grounded search.
Starting from an original human-written RTL implementation, Dr.~RTL searches for functionally equivalent designs
with better timing and area under a fixed commercial
synthesis flow.
The optimization preserves pipeline latency and micro-architectural behavior while allowing combinational
and sequential RTL transformations.\looseness=-1

The framework coordinates three specialized agents:
(1) a timing analysis agent that examines post-synthesis
timing reports, maps critical paths to RTL structures,
and diagnoses bottlenecks;
(2) an RTL optimization agent that generates multiple
candidate transformations in parallel; and
(3) an evaluation agent that runs commercial synthesis
and SEC to obtain PPA and correctness feedback.
An orchestrator coordinates these agents, records
optimization trajectories, and selects the best
equivalent candidate for the next iteration.\looseness=-1

This closed-loop workflow connects optimization decisions
directly to measured implementation quality.
Fine-grained timing analysis identifies specific bottlenecks,
such as deep combinational paths, high-fanout signals,
and complex control--data interactions.
Parallel exploration evaluates different transformations
under the same design constraints, while SEC ensures
that selected implementations preserve functionality.

\textbf{Learning reusable optimization skills.}
Beyond optimizing individual designs, Dr.~RTL learns
reusable knowledge from previous optimization attempts.
Each trajectory records the diagnosed bottleneck,
applied RTL transformation, and resulting PPA and
SEC outcomes.
Because optimization effects depend on the design
and synthesis environment, Dr.~RTL introduces
\emph{group-relative skill learning} to identify
effective transformations under comparable conditions.\looseness=-1

Candidates generated
from the same parent RTL are compared using their
relative optimization outcomes.
A skill-learning agent analyzes these comparisons
and extracts reusable \emph{pattern--strategy} pairs.
Patterns describe recurring bottlenecks, such as
deep decode logic, high fanout, or mux-heavy selection,
while strategies describe corresponding transformations,
such as condition pre-computation, signal replication,
or arithmetic restructuring.

\begin{figure*}[!t]
    \centering
    \vspace{-.3in}
    \includegraphics[width=\linewidth]{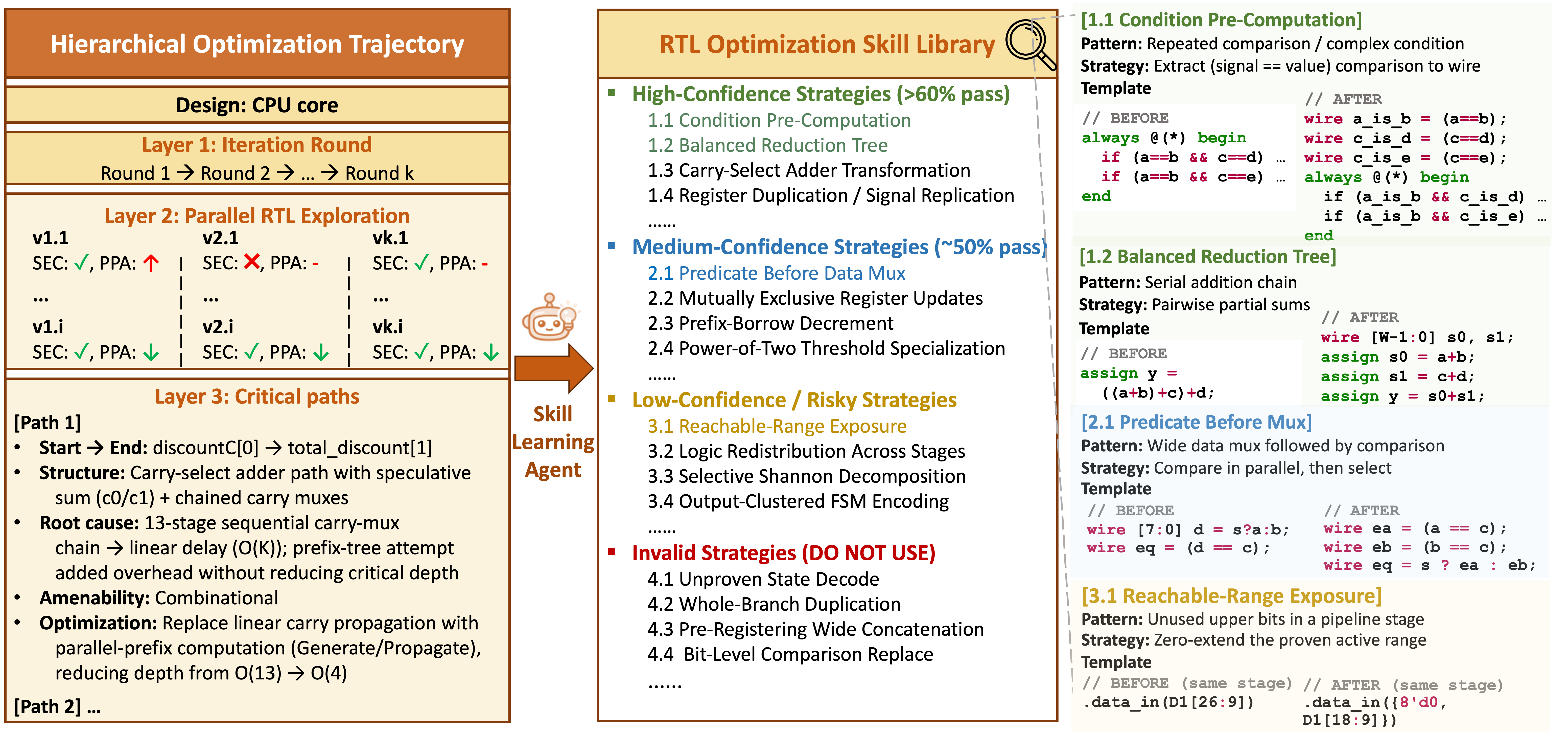}
    \vspace{-.3in}
    \caption{Distilling hierarchical optimization
    trajectories into a reusable skill library.
    Trajectories are organized across iterations,
    parallel candidates, and critical paths,
    supporting the extraction of optimization
    strategies with different confidence levels
    and associated transformation templates.}
    \label{fig:skill-demo}
    \vspace{-.2in}
\end{figure*}

The extracted skills are stored in a confidence-aware
library, with their reliability updated using empirical
PPA improvements and SEC results.
Successful strategies are prioritized for reuse,
while ineffective or invalid transformations are
recorded to avoid repeated exploration.
When optimizing a new design, the agent applies
relevant learned skills or explores new transformations
when no suitable skill is available.
The resulting experience further updates the library,
creating a continual self-improvement loop.

\subsection{Experimental Results}
\label{sec:expr}

\textbf{Experimental setup.}
We evaluate Dr.~RTL on 20 original human-written Verilog
designs covering processors, communication interfaces,
buffers, signal-processing modules, and other digital
systems.
The designs contain 812 lines of RTL on average.
We use Synopsys Design Compiler for commercial synthesis,
the Nangate 45\,nm standard-cell library~\cite{NanGate},
and Cadence JasperGold SEC~\cite{SEC} for verification.
Each design is optimized over 10 iterations with five
parallel candidates per iteration.

\textbf{RTL optimization performance.}
As summarized in~\Cref{tbl:baseline2}, Dr.~RTL improves
timing on all 20 designs, achieving average WNS and TNS
improvements of 21.3\% and 16.9\%, respectively,
with a 5.8\% reduction in area over commercial synthesis.
Nine designs achieve Pareto improvements in timing and
area, and the average SEC pass rate is 86\%.
These results demonstrate substantial RTL optimization
opportunities beyond commercial synthesis, even for
original human-written designs.
We also visualize the optimization trajectory in~\Cref{fig:trajectory}.

Compared with RTLRewriter~\cite{yao2024rtlrewriter}
and SymRTLo~\cite{wang2025symrtlo} under the same
commercial evaluation flow, Dr.~RTL achieves
substantially larger timing improvements.
The iterative baselines use the same backbone LLM
and optimization budget, supporting the benefit
of tool-grounded multi-agent exploration and
reusable optimization knowledge.

\textbf{Skill generalization and interpretability.}
We evaluate cross-design skill transfer through
four-fold cross-validation, learning skills from
15 designs and applying them to five unseen designs
in each fold.
Dr.~RTL extracts 47 pattern--strategy entries:
12 high-confidence, 16 medium-confidence,
six low-confidence, and 13 avoid strategies.
As illustrated in~\Cref{fig:skill-demo}, these
skills connect critical-path patterns with
optimization strategies and their empirical outcomes,
providing interpretable knowledge for future searches.

Overall, Dr.~RTL demonstrates two complementary
capabilities: autonomous RTL optimization through
commercial EDA feedback and continual improvement
through reusable optimization knowledge.
While Dr.~RTL explores alternative implementations
under a fixed synthesis procedure, the next section
extends agentic optimization to the synthesis
procedures themselves.

%% file: section/3-synthesis.tex
\section{Agentic Synthesis Optimization and Algorithm Discovery}
\label{sec:synthesis}

\subsection{Overview}

Beyond RTL optimization, further PPA improvements can be
achieved by optimizing the synthesis procedures themselves.
We investigate two complementary levels of
\emph{design-specific synthesis optimization}:
(1) agentic script tuning, which explores the configuration
space of commercial synthesis tools through tool-grounded
search and reusable optimization skills, and
(2) synthesis algorithm discovery, which develops customized
optimization procedures for individual designs.
These approaches extend agentic optimization from
hardware implementations to the algorithms that produce
them, enabling a broader exploration of the RTL--synthesis
optimization space.

\subsection{Background and Related Work}

Logic synthesis transforms RTL into a technology-mapped
netlist through intermediate representations,
technology-independent optimization, and technology
mapping~\cite{brayton2006scalable,hassoun2012logic}.
Existing research improves these procedures through
technology-aware rewriting~\cite{chen2024syn},
equality saturation~\cite{chen2025morphic,yin2025boole},
learning-guided synthesis~\cite{neto2019lsoracle},
and technology mapping optimization~\cite{neto2021slap}.
We focus on two complementary directions for adapting
synthesis to individual designs.

\textbf{Synthesis flow tuning.}
Flow tuning optimizes the configurations and sequences
of existing synthesis operations.
SynTunSys~\cite{ziegler2016synthesis} automates synthesis
parameter exploration, while FIST~\cite{xie2020fist}
uses feature importance to guide parameter selection.
FlowTune~\cite{neto2022flowtune} explores logic optimization
flows through bandit-based search, and
FlowTuner~\cite{liang2021flowtuner} uses knowledge transfer
for multi-stage EDA flow tuning.
These studies demonstrate the benefits of adapting
synthesis configurations to different designs.
Recent advances in AI agents create further opportunities
for autonomous strategy selection, tool interaction,
and learning reusable optimization knowledge.

\textbf{Synthesis algorithm discovery.}
AlphaEvolve~\cite{novikov2025alphaevolve} demonstrates the
feasibility of AI-driven algorithm discovery through
evaluator-guided code evolution.
Building on this paradigm,
Self-Evolved ABC~\cite{yu2026selfevolvedabc} applies
multi-agent code evolution to improve general-purpose
logic synthesis algorithms.
However, circuit structures and optimization bottlenecks
vary across designs, and general-purpose algorithms
may leave design-specific optimization opportunities unexplored.
With increasingly capable and inexpensive AI coding,
we envision customizing synthesis algorithms for each
individual design to explore its optimization space more
thoroughly and achieve better PPA.

\subsection{Method: Agentic Script Tuning and Algorithm Discovery}
\label{sec:synthesis-method}

\begin{figure}[!t]
    \centering
    \includegraphics[width=\columnwidth]{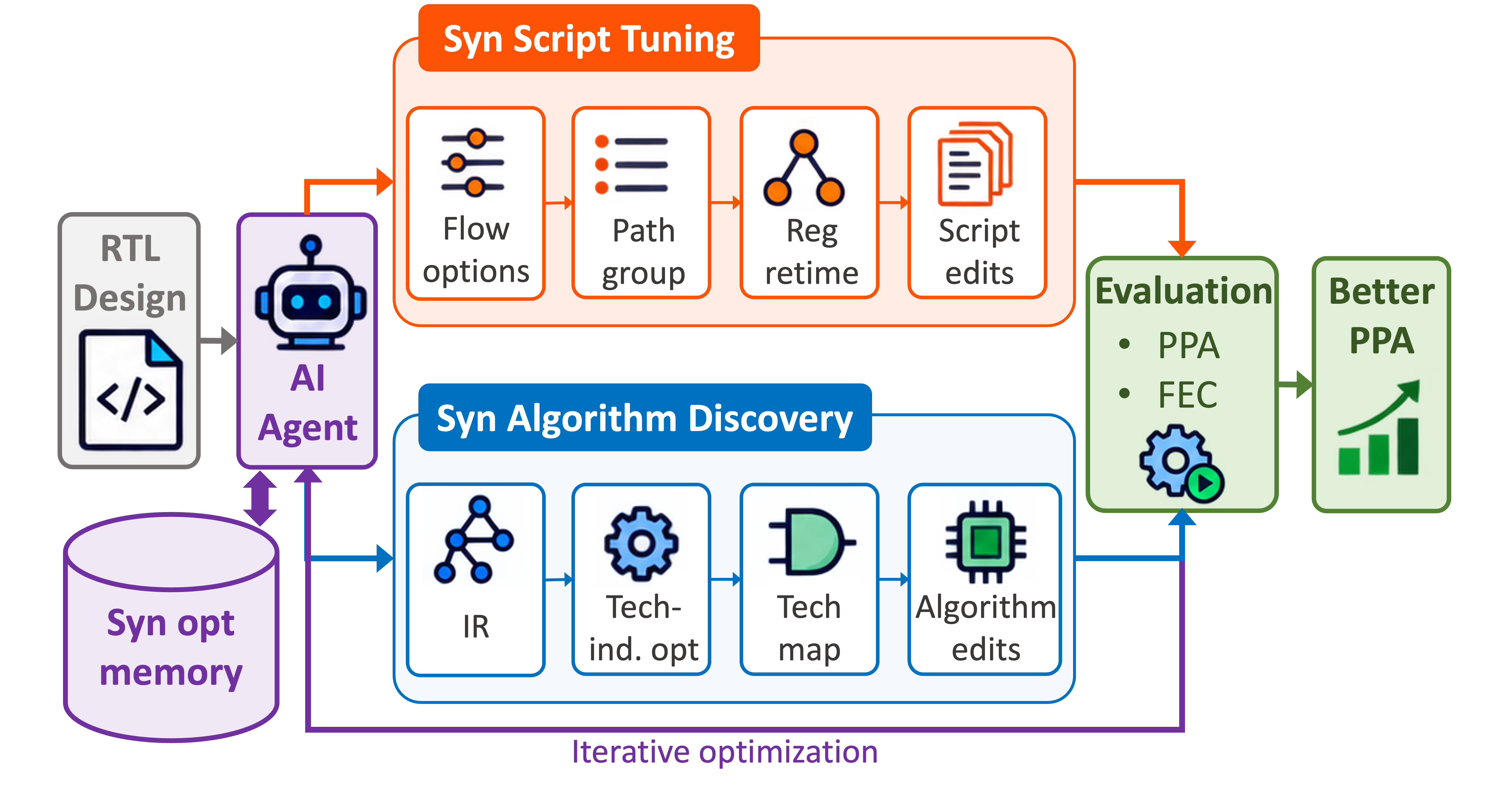}
    \caption{Our two levels of agentic synthesis optimization:
 script tuning for the commercial tool and design-specific
algorithm discovery for the open-source tool.}
    \vspace{-.2in}
    \label{fig:synthesis-method}
\end{figure}

Figure~\ref{fig:synthesis-method} illustrates two levels
of agentic synthesis optimization.
Script tuning explores the configuration space of an
existing synthesis engine, while algorithm discovery
expands the search space by modifying the underlying
optimization procedures.
Both approaches follow a feedback-driven process
under fixed RTL, technology libraries, and design
constraints.

\textbf{Agentic synthesis flow tuning.}
Our first approach extends Dr.~RTL with a synthesis-strategy
agent that explores design-specific configurations of
Synopsys Design Compiler (DC).
Starting from the strong commercial \texttt{compile\_ultra}
flow, the agent analyzes timing bottlenecks, register
connectivity, and previous optimization outcomes to
identify opportunities for improving synthesis scripts.\looseness=-1

The framework follows an agentic workflow of
\emph{analysis, strategy generation, tool-based evaluation,
and skill learning}.
The agent explores alternative synthesis configurations,
evaluates their PPA through commercial synthesis,
and uses the resulting feedback to refine its scripts.
The search considers interactions between different
synthesis operations, allowing optimization effort
to be allocated according to the characteristics
and critical paths of each design.\looseness=-1

To support self-improvement, the agent also accumulates
\emph{synthesis optimization skills}.
These skills capture effective configuration strategies,
their applicable circuit characteristics, and their
observed PPA outcomes.
The accumulated knowledge guides subsequent optimization
attempts, extending Dr.~RTL's skill-based learning
from RTL transformations to synthesis strategies.
This enables agents to systematically explore the
capabilities of commercial synthesis tools while
reusing experience across designs.

\begin{table}[!t]
    \centering
    \caption{Results of (a) commercial DC script tuning
    and (b) open-source synthesis algorithm discovery.
    Baselines are Dr.~RTL's best RTL under
    \texttt{compile\_ultra} and agent-tuned Yosys+ABC,
    respectively. Positive percentages indicate improvement.
    Mean is arithmetic.}
    \vspace{-.2in}
    \label{tab:synthesis-results}
    \small
    \setlength{\tabcolsep}{2pt}
    \begin{tabular*}{\columnwidth}{@{\extracolsep{\fill}}lrrrrrr@{}}
        \multicolumn{7}{@{}l}{\textit{(a) Synthesis script tuning}} \\
        \toprule
        \multirow{3}{*}{Design}
        & \multicolumn{3}{c}{Baseline}
        & \multicolumn{3}{c}{Reduction $\uparrow$} \\
        \cmidrule(lr){2-4}\cmidrule(lr){5-7}
        & WNS & TNS & Area & WNS & TNS & Area \\
        & (ns) & (ns) & ($\mu$m$^2$) & & & \\
        \midrule
        FIFO
        & $-0.45$ & $-1801.81$ & $34987$
        & $7.0\%$ & $2.5\%$ & $0.1\%$ \\
        SHA3
        & $-0.74$ & $-1494.83$ & $33639$
        & $4.1\%$ & $12.9\%$ & $-0.2\%$ \\
        SPI
        & $-0.24$ & $-25.67$ & $1881$
        & $6.9\%$ & $14.8\%$ & $0.5\%$ \\
        UART
        & $-0.35$ & $-22.40$ & $1035$
        & $3.9\%$ & $5.8\%$ & $0.2\%$ \\
        ctrl
        & $-0.35$ & $-548.55$ & $17259$
        & $10.4\%$ & $6.4\%$ & $-7.3\%$ \\
        arm\_cpu1
        & $-2.21$ & $-2002.67$ & $24785$
        & $9.5\%$ & $17.0\%$ & $-1.6\%$ \\
        arm\_cpu2
        & $-0.86$ & $-597.70$ & $13342$
        & $11.8\%$ & $18.9\%$ & $-4.7\%$ \\
        i2c
        & $-0.33$ & $-28.05$ & $1369$
        & $6.1\%$ & $10.2\%$ & $0.7\%$ \\
        \midrule
        Mean
        & $-0.69$ & $-815.21$ & $16037$
        & $7.4\%$ & $11.1\%$ & $-1.6\%$ \\
        \bottomrule
    \end{tabular*}

    \par\medskip

    \begin{tabular*}{\columnwidth}{@{\extracolsep{\fill}}lrrrrr@{}}
        \multicolumn{6}{@{}l}{\textit{(b) Algorithm discovery}} \\
        \toprule
        \multirow{3}{*}{Design}
        & \multicolumn{2}{c}{Baseline}
        & \multicolumn{3}{c}{Reduction $\uparrow$} \\
        \cmidrule(lr){2-3}\cmidrule(lr){4-6}
        & Area & Delay & Area & Delay & ADP \\
        & ($\mu$m$^2$) & (ns) & & & \\
        \midrule
        picorv32
        & $18200$ & $0.56$
        & $12.2\%$ & $2.7\%$ & $14.6\%$ \\
        aes128
        & $15458$ & $0.60$
        & $10.0\%$ & $1.2\%$ & $11.1\%$ \\
        sha256
        & $15075$ & $1.11$
        & $3.6\%$ & $22.8\%$ & $25.6\%$ \\
        spi\_master
        & $2999$ & $0.67$
        & $16.5\%$ & $20.0\%$ & $33.2\%$ \\
        sasc\_uart
        & $958$ & $0.29$
        & $11.1\%$ & $14.3\%$ & $23.8\%$ \\
        usb\_phy
        & $799$ & $0.26$
        & $11.8\%$ & $8.2\%$ & $19.0\%$ \\
        wb\_conmax
        & $26610$ & $0.60$
        & $10.3\%$ & $3.5\%$ & $13.4\%$ \\
        vga\_lcd
        & $129958$ & $1.00$
        & $8.9\%$ & $0.0\%$ & $8.9\%$ \\
        \midrule
        Mean
        & --- & ---
        & $10.6\%$ & $9.1\%$ & $18.7\%$ \\
        \bottomrule
    \end{tabular*}
\end{table}

\textbf{Design-specific synthesis algorithm discovery.}
Our second approach expands the optimization space
from synthesis configurations to synthesis algorithms.
In principle, a synthesis engine capable of finding
the globally optimal implementation within the allowed
design space would eliminate the need for separate RTL
optimization within that same space.
In practice, existing synthesis algorithms are limited
by their circuit representations, available transformations,
search heuristics, and computational budgets.
Consequently, RTL optimization can expose better
implementations that existing synthesis procedures
fail to discover directly.

This observation motivates a broader direction:
\emph{improving the synthesis algorithms themselves}.
As AI coding becomes increasingly powerful and inexpensive, we can
trade additional optimization time for better PPA
by developing a customized synthesizer for each design.
Rather than requiring a single algorithm to work well
across all circuits, agents can specialize synthesis
procedures to exploit the structural characteristics
and optimization opportunities of the target design.
The objective is to explore the available implementation
space more thoroughly and approach better design-specific
PPA solutions.

We explore algorithm discovery across three synthesis
stages, as illustrated in~\Cref{fig:synthesis-method}:

\begin{itemize}
    \item \emph{Intermediate representation (IR).}
    Agents explore circuit representations and
    decomposition procedures that expose design-specific
    structures and optimization opportunities.

    \item \emph{Technology-independent optimization.}
    Agents develop rewriting, factoring, restructuring,
    and logic-sharing algorithms tailored to the
    target circuit.

    \item \emph{Technology mapping.}
    Agents adapt functional matching, covering,
    and cell-selection algorithms to improve area
    and timing under the target technology library.
\end{itemize}

The three stages provide complementary optimization
freedom, from choosing circuit representations to
transforming logic structures and selecting physical
cells.
Exploring them together allows agents to discover
optimization procedures that are particularly
effective for individual designs.

Algorithm discovery follows a verification-guided
code evolution loop.
The agent modifies executable synthesis code,
regenerates a netlist from the original RTL,
checks functional equivalence, and evaluates
area and delay.
Critical-path and implementation feedback guide
subsequent algorithm revisions, while successful
program variants are retained.
The resulting customized synthesizer can
reproduce the optimized implementation from
the original RTL.

\subsection{Evaluation Results}
\label{sec:synthesis-results}

We evaluate the two approaches on eight designs each,
using commercial and open-source synthesis environments,
respectively.
Script tuning targets Synopsys Design Compiler,
while algorithm discovery uses Yosys+ABC.
Both experiments use fixed RTL, design constraints,
and the Nangate 45\,nm technology library.

\textbf{Commercial synthesis script tuning.}
We evaluate script tuning on RTL already optimized by
Dr.~RTL, using DC \texttt{compile\_ultra} as the baseline.
The agent explores path grouping through
\texttt{group\_path} and retiming through
\texttt{optimize\_registers}, combining four strategies
for each in a $4\times4$ search matrix.
We perform five additional search rounds using
WNS, TNS, and area feedback.

As shown in~\Cref{tab:synthesis-results}(a),
script tuning improves WNS and TNS on all eight designs,
with average improvements of 7.4\% and 11.1\%,
respectively, at a 1.6\% average area overhead.
These results demonstrate further timing optimization
opportunities beyond Dr.~RTL under commercial synthesis.

\textbf{Open-source synthesis algorithm discovery.}
We compare design-specific algorithm evolution with
agent-tuned Yosys+ABC.
Each evolved synthesis program must generate a
functionally equivalent netlist from the original RTL.

As shown in~\Cref{tab:synthesis-results}(b),
algorithm discovery achieves average reductions of
10.6\% in area, 9.1\% in delay, and 18.7\% in
area--delay product (ADP), with seven of eight designs
improving both area and delay.
These results demonstrate the potential of
design-specific algorithms to improve PPA beyond
agent-tuned general-purpose synthesis flows.

Together, these experiments show the benefits of
agentic optimization at both the synthesis configuration
and algorithmic levels, motivating joint exploration
of RTL implementations and EDA optimization procedures.

%% file: section/5-discussion.tex
\section{Future Vision: Toward Verification-Guided Agentic DSE}
\label{sec:future}

\subsection{Hierarchical Optimization Spaces}

\begin{figure}[!t]
    \centering
    \includegraphics[width=\columnwidth]{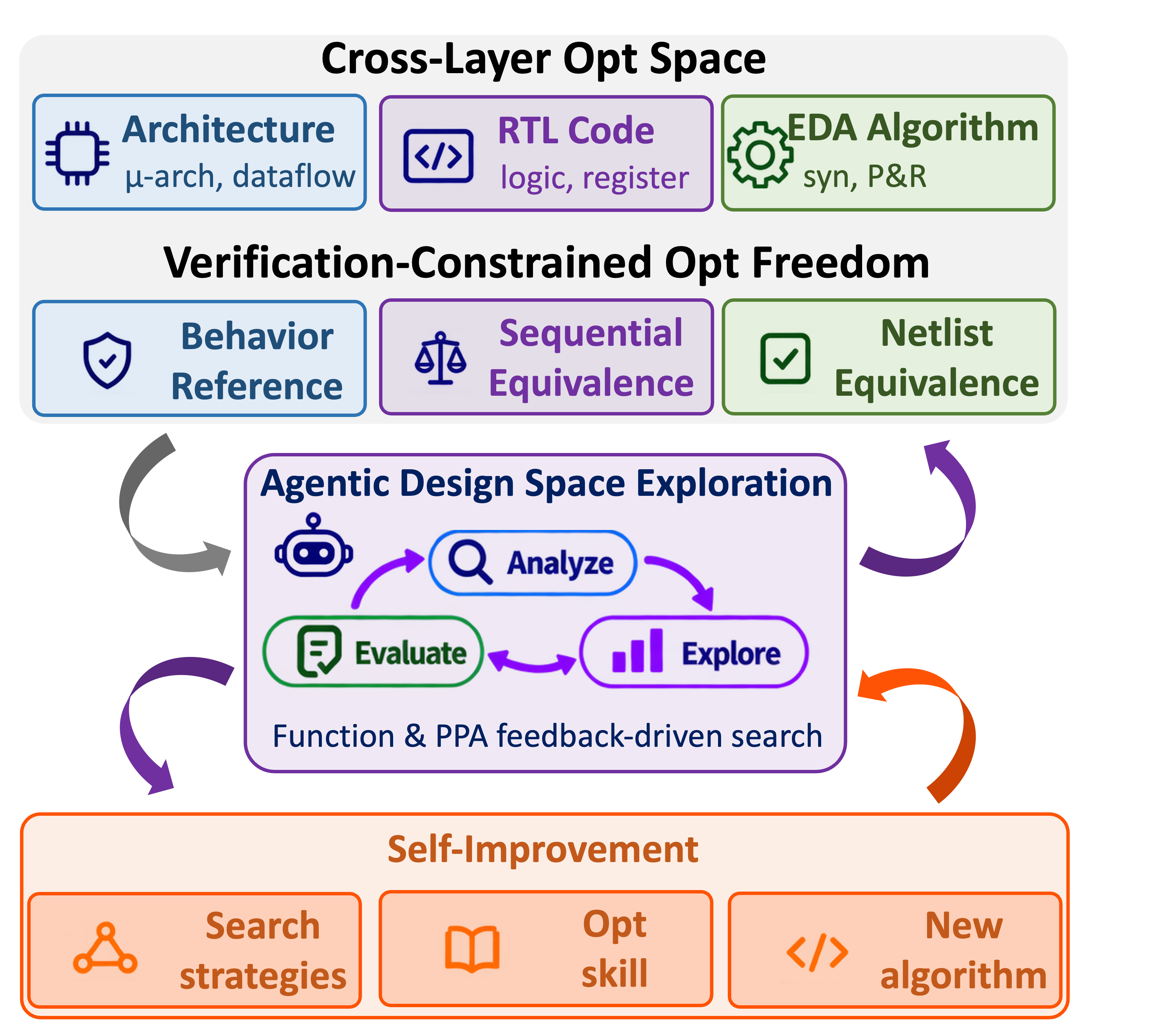}
    \vspace{-.2in}
    \caption{Future vision of self-improving cross-layer
    agentic DSE. Optimization spans architecture, RTL,
    and EDA algorithms, with verification defining
    the allowable freedom at each level.
    Tool feedback and reusable knowledge support
    continual improvement across design stages.}
    \vspace{-.1in}
    \label{fig:future}
\end{figure}

Existing AI-driven hardware optimization methods largely
focus on individual design stages, such as architectural
exploration, RTL optimization, or synthesis tool optimization.
However, these stages are closely coupled, and optimizing
them independently limits the available design space.
We envision a broader cross-layer optimization space
organized across three levels: architecture, RTL implementation,
and EDA algorithms, as illustrated in~\Cref{fig:future}.
At the architectural level, the design space covers
microarchitectures, dataflows, and resource organizations
~\cite{bai2021boom,bai2023archexplorer}.
At the RTL level, it includes logic restructuring,
resource sharing, and sequential transformations.
At the EDA tool level, it covers synthesis and physical
design configurations, transformations, and algorithms.

These spaces interact across abstraction levels:
architectural decisions shape the RTL implementation space,
while synthesis and physical design procedures influence
the achievable PPA of different implementations.
Joint exploration can therefore discover optimization
opportunities that are difficult to identify when
each stage is optimized independently.
Clearly defining these spaces and their relationships
provides a foundation for cross-layer agentic DSE.\looseness=-1

\subsection{Verification-Constrained Optimization Freedom}

Cross-layer optimization requires a clear definition
of what must be preserved and what can be changed
at each level.
We envision verification constraints that separate
\emph{verified invariants} from \emph{optimization freedom}.
Architectural exploration preserves functional behavior
and interface requirements, RTL optimization preserves
sequential behavior, and EDA algorithm optimization
ensures netlist correctness and implementation constraints.
These requirements can be checked through simulation,
equivalence checking, or other verification techniques.

Importantly, cross-layer exploration may involve
coordinated changes across multiple design stages.
Explicit verification constraints define which changes
are valid and help maintain correctness as optimization
moves between architecture, RTL, and EDA algorithms.
This enables agents to explore a broader joint
optimization space with reliable correctness feedback.

\subsection{Toward Self-Improving Agentic Design-Space Exploration}

Moving from single-stage optimization to cross-layer
DSE introduces a substantially larger and more
interconnected search space.
AI agents offer opportunities to coordinate optimization
across abstraction levels using circuit characteristics,
implementation feedback, and accumulated experience.
However, effective exploration requires deciding
which design levels to optimize, how to coordinate
transformations, and how to allocate search effort
across the joint space.

This raises an important research question:
\emph{How can AI agents continuously improve their
DSE capabilities across interconnected hardware
design spaces?}
Future research can explore adaptive cross-layer
optimization strategies and the reuse of knowledge
learned from previous searches, including effective
transformations, search policies, and optimization
algorithms.
Ultimately, we envision self-improving AI-native EDA
systems that jointly optimize hardware architectures,
RTL implementations, and the EDA procedures used
to realize them.

%% file: section/6-conclusion.tex
\section{Conclusion}
\label{sec:concl}

This paper presents three connected directions toward
autonomous RTL and synthesis optimization: realistic
benchmarking, agentic
tool-grounded RTL optimization, and
design-specific synthesis optimization.
We introduce complementary benchmarking settings to
evaluate meaningful PPA improvements beyond synthesis,
and present Dr.~RTL, a self-improving agentic framework
that achieves substantial timing and area improvements
over commercial synthesis.
We further demonstrate the potential of agent-driven
synthesis flow tuning and design-specific algorithm
discovery for additional PPA gains.
Looking ahead, we envision verification-guided agentic
DSE across architecture, RTL, and EDA algorithms.
By organizing optimization spaces, defining verification
constraints, and accumulating reusable optimization
knowledge, future agents can progressively improve
both hardware implementations and the EDA procedures
that realize them.

\section{ACKNOWLEDGEMENT}
This work is supported by Hong Kong Research Grants Council (RGC) CRF-YCRG C6003-24Y, 16200724, T46-415/25-R, and JLFS-YSF/E-605/26. It was partially conducted by ACCESS – AI Chip Center for Emerging Smart Systems, supported by the InnoHK initiative of the Innovation and Technology Commission of the Hong Kong Special Administrative Region Government.

%% file: ref.bib
@inproceedings{chen2024syn,
  title={E-syn: E-graph rewriting with technology-aware cost functions for logic synthesis},
  author={Chen, Chen and Hu, Guangyu and Zuo, Dongsheng and Yu, Cunxi and Ma, Yuzhe and Zhang, Hongce},
  booktitle={ Design Automation Conference (DAC)},
  year={2024}
}

@inproceedings{bai2021boom,
 author = {Bai, Chen and Sun, Qi and Zhai, Jianwang and Ma, Yuzhe and Yu, Bei and Wong, Martin DF},
 booktitle={ International Conference on Computer-Aided Design (ICCAD) },
 title = {BOOM-Explorer: RISC-V BOOM microarchitecture design space exploration framework},
 year = {2021}
}

@inproceedings{bai2023archexplorer,
 author = {Bai, Chen and Huang, Jiayi and Wei, Xuechao and Ma, Yuzhe and Li, Sicheng and Zheng, Hongzhong and Yu, Bei and Xie, Yuan},
 booktitle={ International Symposium on Microarchitecture (MICRO) },
 title = {ArchExplorer: Microarchitecture exploration via bottleneck analysis},
 year = {2023}
}

@inproceedings{brayton2006scalable,
 author = {Brayton, Alan and Mishchenko, Robert and Mishchenko, A},
 booktitle={ International Workshop on Logic and Synthesis (IWLS) },
 title = {Scalable logic synthesis using a simple circuit structure},
 year = {2006}
}

@inproceedings{fang2024annotating,
 author = {Fang, Wenji and Liu, Shang and Zhang, Hongce and Xie, Zhiyao},
 booktitle={ Design Automation Conference (DAC) },
 title = {Annotating Slack Directly on Your Verilog: Fine-Grained RTL Timing Evaluation for Early Optimization},
 year = {2024}
}

@inproceedings{liang2021flowtuner,
 author = {Liang, Rongjian and Jung, Jinwook and Xiang, Hua and Reddy, Lakshmi and Lvov, Alexey and Hu, Jiang and Nam, Gi-Joon},
 booktitle={ International Conference on Computer-Aided Design (ICCAD) },
 title = {FlowTuner: A multi-stage EDA flow tuner exploiting parameter knowledge transfer},
 year = {2021}
}

@article{liu2023verilogeval,
 author = {Liu, Mingjie and Pinckney, Nathaniel and Khailany, Brucek and Ren, Haoxing},
 journal={ arXiv preprint arXiv:2309.07544 },
 title = {VerilogEval: Evaluating Large Language Models for Verilog Code Generation},
 year = {2023}
}

@article{liu2024rtlcoder,
 author = {Liu, Shang and Fang, Wenji and Lu, Yao and Zhang, Qijun and Zhang, Hongce and Xie, Zhiyao},
 journal={ IEEE Transactions on Computer-Aided Design of Integrated Circuits and Systems (TCAD) },
 title = {RTLCoder: Fully Open-Source and Efficient LLM-Assisted RTL Code Generation Technique},
 year = {2024}
}

@inproceedings{lu2024rtllm,
 author = {Lu, Yao and Liu, Shang and Zhang, Qijun and Xie, Zhiyao},
 booktitle={ Asia and South Pacific Design Automation Conference (ASP-DAC) },
 title = {{RTLLM}: An Open-Source Benchmark for Design RTL Generation with Large Language Model},
 year = {2024}
}

@misc{NanGate,
 author = {Si2},
 title = {{NanGate} 45nm Open Cell Library},
 year = {2018}
}

@inproceedings{neto2019lsoracle,
 author = {Neto, Walter Lau and Austin, Max and Temple, Scott and Amaru, Luca and Tang, Xifan and Gaillardon, Pierre-Emmanuel},
 booktitle={ International Conference on Computer-Aided Design (ICCAD) },
 title = {{LSOracle}: A logic synthesis framework driven by artificial intelligence},
 year = {2019}
}

@inproceedings{neto2021slap,
 author = {Neto, Walter Lau and Moreira, Matheus T and Li, Yingjie and Amar{\`u}, Luca and Yu, Cunxi and Gaillardon, Pierre-Emmanuel},
 booktitle={ Design Automation Conference (DAC) },
 title = {{SLAP}: A Supervised Learning Approach for Priority Cuts Technology Mapping},
 year = {2021}
}

@article{coward2024rover,
  title={Rover: Rtl optimization via verified e-graph rewriting},
  author={Coward, Samuel and Drane, Theo and Constantinides, George A},
  journal={IEEE Transactions on Computer-Aided Design of Integrated Circuits and Systems},
  year={2024},
  publisher={IEEE}
}

@inproceedings{xie2020fist,
 author = {Xie, Zhiyao and Fang, Guan-Qi and Huang, Yu-Hung and Ren, Haoxing and Zhang, Yanqing and Khailany, Brucek and Fang, Shao-Yun and Hu, Jiang and Chen, Yiran and Barboza, Erick Carvajal},
 booktitle={ Asia and South Pacific Design Automation Conference (ASP-DAC) },
 title = {{FIST}: A feature-importance sampling and tree-based method for automatic design flow parameter tuning},
 year = {2020}
}

@article{neto2022flowtune,
  title={Flowtune: End-to-end automatic logic optimization exploration via domain-specific multiarmed bandit},
  author={Neto, Walter Lau and Li, Yingjie and Gaillardon, Pierre-Emmanuel and Yu, Cunxi},
  journal={IEEE Transactions on Computer-Aided Design of Integrated Circuits and Systems (TCAD)},
  year={2022},
}

@article{thorat2023advanced,
  title={Advanced Large Language Model (LLM)-Driven Verilog Development: Enhancing Power, Performance, and Area Optimization in Code Synthesis},
  author={Thorat, Kiran and Zhao, Jiahui and Liu, Yaotian and Peng, Hongwu and Xie, Xi and Lei, Bin and Zhang, Jeff and Ding, Caiwen},
  journal={arXiv preprint arXiv:2312.01022},
  year={2023}
}

@article{fang2025survey,
  title={A survey of circuit foundation model: Foundation ai models for vlsi circuit design and eda},
  author={Fang, Wenji and Wang, Jing and Lu, Yao and Liu, Shang and Wu, Yuchao and Ma, Yuzhe and Xie, Zhiyao},
  journal={arXiv preprint arXiv:2504.03711},
  year={2025}
}

@article{chen2025morphic,
  title={E-morphic: Scalable Equality Saturation for Structural Exploration in Logic Synthesis},
  author={Chen, Chen and Hu, Guangyu and Yu, Cunxi and Ma, Yuzhe and Zhang, Hongce},
  journal={arXiv preprint arXiv:2504.11574},
  year={2025}
}

@inproceedings{yao2024rtlrewriter,
  title={Rtlrewriter: Methodologies for large models aided rtl code optimization},
  author={Yao, Xufeng and Wang, Yiwen and Li, Xing and Lian, Yingzhao and Chen, Ran and Chen, Lei and Yuan, Mingxuan and Xu, Hong and Yu, Bei},
  booktitle={Proceedings of IEEE/ACM International Conference on Computer-Aided Design (ICCAD)},
  year={2024}
}

@inproceedings{wang2025symrtlo,
 title={Symrtlo: Enhancing rtl code optimization with llms and neuron-inspired symbolic reasoning},
 author={Wang, Yiting and Ye, Wanghao and Guo, Ping and He, Yexiao and Wang, Ziyao and Tian, Bowei and He, Shwai and Sun, Guoheng and Shen, Zheyu and Chen, Sihan and others},
 booktitle={ Advances in Neural Information Processing Systems (NeurIPS) },
 year={2025}
}

@article{pinckney2025comprehensive,
  title={Comprehensive Verilog design problems: A next-generation benchmark dataset for evaluating large language models and agents on rtl design and verification},
  author={Pinckney, Nathaniel and Deng, Chenhui and Ho, Chia-Tung and Tsai, Yun-Da and Liu, Mingjie and Zhou, Wenfei and Khailany, Brucek and Ren, Haoxing},
  journal={arXiv preprint arXiv:2506.14074},
  year={2025}
}

@article{xu2025rethinking,
  title={Rethinking LLM-Based RTL Code Optimization Via Timing Logic Metamorphosis},
  author={Xu, Zhihao and Li, Bixin and Wang, Lulu},
  journal={arXiv preprint arXiv:2507.16808},
  year={2025}
}

@inproceedings{thorat2025llm,
  title={LLM-VeriPPA: Power, Performance, and Area Optimization aware Verilog Code Generation with Large Language Models},
  author={Thorat, Kiran and Zhao, Jiahui and Liu, Yaotian and Hasan, Amit and Peng, Hongwu and Xie, Xi and Lei, Bin and Ding, Caiwen},
  booktitle={2025 ACM/IEEE 7th Symposium on Machine Learning for CAD (MLCAD)},
  pages={1--7},
  year={2025},
  organization={IEEE}
}

@inproceedings{tasnia2025veriopt,
  title={Veriopt: Ppa-aware high-quality verilog generation via multi-role llms},
  author={Tasnia, Kimia and Garcia, Alexander and Farheen, Tasnuva and Rahman, Sazadur},
  booktitle={2025 IEEE/ACM International Conference On Computer Aided Design (ICCAD)},
  pages={1--9},
  year={2025},
  organization={IEEE}
}

@article{lu2026new,
  title={A New Benchmark for the Appropriate Evaluation of RTL Code Optimization},
  author={Lu, Yao and Liu, Shang and Zhou, Hangan and Fang, Wenji and Zhang, Qijun and Xie, Zhiyao},
  journal={arXiv preprint arXiv:2601.01765},
  year={2026}
}

@article{hsin2026evolve,
  title={EvolVE: Evolutionary Search for LLM-based Verilog Generation and Optimization},
  author={Hsin, Wei-Po and Deng, Ren-Hao and Hsieh, Yao-Ting and Huang, En-Ming and Hung, Shih-Hao},
  journal={arXiv preprint arXiv:2601.18067},
  year={2026}
}

@inproceedings{chang2026codmas,
  title={CODMAS: A Dialectic Multi-Agent Collaborative Framework for Structured RTL Optimization},
  author={Chang, Che-Ming and Vijayaraghavan, Prashanth and Jadhav, Ashutosh and Mackin, Charles and Tsai, Hsinyu and Mukherjee, Vandana and Degan, Ehsan},
  booktitle={Proceedings of the 19th Conference of the European Chapter of the Association for Computational Linguistics (Volume 5: Industry Track)},
  pages={777--788},
  year={2026}
}

@article{xia2026skillrl,
  title={SkillRL: Evolving Agents via Recursive Skill-Augmented Reinforcement Learning},
  author={Xia, Peng and Chen, Jianwen and Wang, Hanyang and Liu, Jiaqi and Zeng, Kaide and Wang, Yu and Han, Siwei and Zhou, Yiyang and Zhao, Xujiang and Chen, Haifeng and others},
  journal={arXiv preprint arXiv:2602.08234},
  year={2026}
}

@article{xu2025mem,
  title={A-mem: Agentic memory for llm agents},
  author={Xu, Wujiang and Liang, Zujie and Mei, Kai and Gao, Hang and Tan, Juntao and Zhang, Yongfeng},
  journal={arXiv preprint arXiv:2502.12110},
  year={2025}
}

@book{vahid2010digital,
  title={Digital design with RTL design, VHDL, and Verilog},
  author={Vahid, Frank},
  year={2010},
  publisher={John Wiley \& Sons}
}

@article{ping2026poet,
  title={POET: Power-Oriented Evolutionary Tuning for LLM-Based RTL PPA Optimization},
  author={Ping, Heng and Zhang, Peiyu and Wang, Zhenkun and Li, Shixuan and Cheng, Anzhe and Yang, Wei and Bogdan, Paul and Nazarian, Shahin},
  journal={arXiv preprint arXiv:2603.19333},
  year={2026}
}

@misc{SEC,
 author = {Cadence Design Systems},
 title = {Cadence JasperGold Sequential Equivalence Checking App},
 howpublished = {\url{https://www.cadence.com/en_US/home/tools/system-design-and-verification/formal-and-static-verification/jasper-verification-platform/jaspergold-sequential-equivalence-checking-app.html}},
  year         = {2026},
}

@book{hassoun2012logic,
  title={Logic synthesis and verification},
  author={Hassoun, Soha and Sasao, Tsutomu},
  volume={654},
  year={2012},
  publisher={Springer Science \& Business Media}
}

@inproceedings{yin2025boole,
  title={Boole: Exact symbolic reasoning via boolean equality saturation},
  author={Yin, Jiaqi and Song, Zhan and Chen, Chen and Hu, Qihao and Yu, Cunxi},
  booktitle={2025 62nd ACM/IEEE Design Automation Conference (DAC)},
  pages={1--7},
  year={2025},
  organization={IEEE}
}

@inproceedings{chowdhury2024retrieval,
  title={Retrieval-guided reinforcement learning for boolean circuit minimization},
  author={Chowdhury, Animesh Basak and Romanelli, Marco and Tan, Benjamin and Karri, Ramesh and Garg, Siddharth},
  booktitle={International Conference on Learning Representations (ICLR)},
  year={2024},
}

@inproceedings{ziegler2016synthesis,
  title={A synthesis-parameter tuning system for autonomous design-space exploration},
  author={Ziegler, Matthew M and Liu, Hung-Yi and Gristede, George and Owens, Bruce and Nigaglioni, Ricardo and Carloni, Luca P},
  booktitle={2016 Design, Automation \& Test in Europe Conference \& Exhibition (DATE)},
  pages={1148--1151},
  year={2016},
  organization={IEEE}
}

@article{sami2024aivril,
  title={Aivril: Ai-driven rtl generation with verification in-the-loop},
  author={Sami, Humza and Gaillardon, Pierre-Emmanuel and Tenace, Valerio and others},
  journal={arXiv preprint arXiv:2409.11411},
  year={2024}
}

@inproceedings{zhao2025mage,
  title={Mage: A multi-agent engine for automated rtl code generation},
  author={Zhao, Yujie and Zhang, Hejia and Huang, Hanxian and Yu, Zhongming and Zhao, Jishen},
  booktitle={2025 62nd ACM/IEEE Design Automation Conference (DAC)},
  pages={1--7},
  year={2025},
  organization={IEEE}
}

@inproceedings{deng2025scalertl,
  title={Scalertl: Scaling llms with reasoning data and test-time compute for accurate rtl code generation},
  author={Deng, Chenhui and Tsai, Yun-Da and Liu, Guan-Ting and Yu, Zhongzhi and Ren, Haoxing},
  booktitle={2025 ACM/IEEE 7th Symposium on Machine Learning for CAD (MLCAD)},
  pages={1--9},
  year={2025},
  organization={IEEE}
}

@article{deng2026ace,
  title={ACE-RTL: When Agentic Context Evolution Meets RTL-Specialized LLMs},
  author={Deng, Chenhui and Yu, Zhongzhi and Liu, Guan-Ting and Pinckney, Nathaniel and Ren, Haoxing},
  journal={arXiv preprint arXiv:2602.10218},
  year={2026}
}

@article{zhang2026rtlseek,
  title={RTLSeek: Boosting the LLM-Based RTL Generation with Multi-Stage Diversity-Oriented Reinforcement Learning},
  author={Zhang, Xinyu and Chao, Zhiteng and Wang, Yonghao and Sun, Bin and Ma, Tianyun and Yang, Tianmeng and Mu, Jianan and Ye, Jing Justin and Li, Huawei},
  journal={arXiv preprint arXiv:2603.27630},
  year={2026}
}

@inproceedings{liu2024openllmrtl,
  author={Liu, Shang and Lu, Yao and Fang, Wenji and Li, Mengming and Xie, Zhiyao},
  title={{OpenLLM-RTL}: Open Dataset and Benchmark for {LLM}-Aided Design {RTL} Generation},
  booktitle={International Conference on Computer-Aided Design (ICCAD)},
  year={2024}
}

@inproceedings{xxx,
  title={{Dr. RTL}: Autonomous Agentic {RTL} Optimization through Tool-Grounded Self-Improvement},
  author={Fang, Wenji and Lu, Yao and Liu, Shang and Wang, Jing and Guo, Ziyan and He, Junxian and Tu, Fengbin and Xie, Zhiyao},
  booktitle={International Conference on Computer-Aided Design (ICCAD)},
  year={2026}
}

@article{novikov2025alphaevolve,
  title={{AlphaEvolve}: A Coding Agent for Scientific and Algorithmic Discovery},
  author={Novikov, Alexander and V{\~u}, Ng{\^a}n and Eisenberger, Marvin and Dupont, Emilien and Huang, Po-Sen and Wagner, Adam Zsolt and Shirobokov, Sergey and Kozlovskii, Borislav and Ruiz, Francisco J. R. and Mehrabian, Abbas and Kumar, M. Pawan and See, Abigail and Chaudhuri, Swarat and Holland, George and Davies, Alex and Nowozin, Sebastian and Kohli, Pushmeet and Balog, Matej},
  journal={arXiv preprint arXiv:2506.13131},
  year={2025},
  doi={10.48550/arXiv.2506.13131}
}

@inproceedings{yu2026selfevolvedabc,
  author={Yu, Cunxi and Ren, Haoxing},
  title={Autonomous Evolution of {EDA} Tools: Multi-Agent Self-Evolved {ABC}},
  booktitle={Proceedings of the 63rd ACM/IEEE Design Automation Conference},
  year={2026},
  doi={10.1145/3770743.3804221}
}
